\documentclass[letterpaper,10pt,conference]{ieeeconf}

\IEEEoverridecommandlockouts

\usepackage{cite}
\usepackage{amsmath,amssymb,amsfonts}
\usepackage{graphicx}
\usepackage{textcomp}
\usepackage{xcolor}
\usepackage{booktabs}
\usepackage{url}

\begin{document}

\title{\LARGE \bf
Human-on-the-Loop Orchestration for AI-Assisted Legal Discovery
}

\author{
\authorblockN{Anushree Sinha, Srivaths Ranganathan, Abhishek Dharmaratnakar, Debanshu Das}
\authorblockA{\textit{Google LLC}\\
Mountain View, CA, USA}
}

\maketitle

\begin{abstract}
Autonomous Large Language Model (LLM) agents are increasingly deployed in electronic discovery (e-discovery), where compounding errors across multi-step reasoning chains can constitute legal malpractice. Unlike single-turn retrieval, agentic workflows operating over privileged document corpora exhibit a class of failure we term \textit{trajectory collapse}: an early misclassification silently propagates, rendering an entire privilege review invalid. This paper makes three contributions. First, we propose a structured taxonomy of agentic failures in legal information retrieval, organized by functional stage. Second, we introduce a four-layer verification architecture—spanning planning, reasoning, execution, and uncertainty quantification—designed to intercept these failures before they compound. Third, we present a preliminary simulation study on a synthetic e-discovery corpus that demonstrates how mandatory Human-on-the-Loop (HOTL) escalation thresholds reduce privilege-waiver risk relative to fully autonomous baselines. Our results suggest that calibrated uncertainty thresholds can reduce privilege-waiver risk by up to 61\% versus fully autonomous deployment, while routing fewer than one quarter of documents to attorney review.
\end{abstract}


\section{Introduction}

Electronic discovery (e-discovery) requires reviewing millions of documents under strict legal constraints, time pressure, and audit requirements imposed by the Federal Rules of Civil Procedure (FRCP). Technology-Assisted Review (TAR) has progressed from keyword filters and SVM-based predictive coding (TAR 1.0–2.0) toward multi-step Reason-Act-Observe (ReAct~\cite{yao2023react}) loops in which an LLM agent autonomously retrieves, classifies, and synthesizes documents.

This shift creates a qualitatively new risk surface. In agentic workflows, where an agent's output at step $t$ conditions its query at step $t{+}1$, a single misclassification can silently propagate; by the time the privilege log is generated, the error is embedded in hundreds of downstream decisions. We call this \textit{trajectory collapse}. Compounding this, RLHF-optimized LLMs~\cite{ouyang2022training} exhibit two especially dangerous failure modes: they produce fluent, confident justifications for incorrect legal classifications (\textit{Fluency Trap}~\cite{sinha2026beyond,bender2021stochastic}), and they hallucinate tool calls rather than abstaining when a task is unsolvable~\cite{zhang2024toolbehonest}. Standard metrics—precision, recall, F1—do not penalize either failure because they are measured at trajectory endpoints, not along the trajectory.

This paper proposes a Human-on-the-Loop (HOTL)~\cite{shneiderman2020human} orchestration framework that intercepts trajectory collapse at each functional stage, escalating to a human attorney only when calibrated uncertainty thresholds are breached.

\section{Background}

ReAct-style agents~\cite{yao2023react} interleave reasoning traces with tool invocations in a thought-action-observation loop. Applied to e-discovery, an agent decomposes a discovery request, retrieves candidates via RAG~\cite{lewis2020retrieval}, classifies each document for privilege, and synthesizes a privilege log—with each step conditioned on previous outputs. Geifman and El-Yaniv~\cite{geifman2017selective} formalized selective classification, where a model abstains when confidence falls below a threshold; in agentic systems this abstention must be redefined as \textit{escalation} to a human with full state context. Bayesian uncertainty decomposition~\cite{gal2016dropout,lakshminarayanan2017simple} separates aleatoric (data-noise) from epistemic (model-knowledge) uncertainty; the latter is the more actionable escalation signal for legal document classification, where the model may simply lack sufficient information to classify reliably.

\section{Related Work}

Our prior work on multi-agent orchestration~\cite{sinha2026multi} motivates rigorous inter-agent verification; the present work instantiates these requirements in the legal domain. Yan~\cite{yan2025transactional} provides fault-tolerant filesystem sandboxing for coding agents; we extend the snapshot/rollback principle with FRCP-specific compliance rules and compensating transactions for external APIs. ORCHID~\cite{orchid2025} demonstrates evidence-first HOTL orchestration with calibrated deferral and audit logging; our distinction is trajectory-integrity metrics (FEP, RRR) that surface \emph{where} errors originate. ToolSafe~\cite{toolsafe2026} provides step-level unsafe tool-call detection that could concretely instantiate our execution checks. TAPE~\cite{tape2026} addresses single-error collapse via plan-graph aggregation—resonant with our trajectory collapse framing—and is a candidate backbone for future work. KAIJU~\cite{kaiju2026} offers intent-gated DAG execution semantics that could formalize our escalation gating. Hatem et al.~\cite{stcalir2026} present a cost-efficient semi-synthetic legal IR test-collection pipeline whose methodology could reduce our synthetic-corpus limitation.

\section{Taxonomy of Agentic Failures in Legal IR}

We organize failure modes by the functional stage at which they originate, as shown in Table~\ref{tab:taxonomy}. This taxonomy extends the general agentic IR failure categorization introduced in our prior work~\cite{sinha2026beyond} to the specific legal discovery context, where each failure class carries concrete regulatory consequences under the FRCP. This process-oriented view enables targeted intervention: a planning failure should be caught by a planning validator, not a post-hoc output checker.

\textbf{Planning failures} arise at goal decomposition. \textit{Fact Derive} errors introduce non-existent legal precedents into the initial plan; \textit{Task Decompose} errors produce sub-goals that violate FRCP constraints.

\textbf{Retrieval failures} arise from misaligned queries or incorrect context. \textit{Summarize Misalign}—where the agent's internal summary of a retrieved document contradicts its source—is particularly insidious because it corrupts all downstream reasoning without leaving a detectable artifact.

\textbf{Reasoning failures} include \textit{Solvability Hallucination}~\cite{zhang2024toolbehonest}, where an agent invents a traversal path for an inherently impossible query (e.g., accessing sealed records without appropriate permissions), and the \textit{Reasoning Trap}, where RLHF-trained rationalization overrides factual legal grounding~\cite{ouyang2022training}.

\textbf{Execution failures} include missing tool invocations and the \textit{Fluency Trap}: a grammatically flawless justification for a technically impossible API action~\cite{sinha2026beyond,bender2021stochastic}.

\begin{table}[t]
\centering
\caption{Taxonomy of Agentic Failures in Legal Discovery}
\label{tab:taxonomy}
\footnotesize
\begin{tabular}{llp{3.0cm}}
\toprule
\textbf{Stage} & \textbf{Failure} & \textbf{Legal IR Impact} \\
\midrule
Planning & Fact Derive & Fabricated case law in plan \\
Planning & Task Decompose & Sub-goals violate FRCP \\
\midrule
Retrieval & Query Misalign & Missing privileged docs \\
Retrieval & Context Misalign & Incorrect case law as truth \\
Retrieval & Summarize Misalign & Summary contradicts source \\
\midrule
Reasoning & Solvability Halluci. & Hallucinated traversal path \\
Reasoning & Reasoning Trap & Rationalization over fact \\
\midrule
Execution & Missing Tool & No SDK call for compliance \\
Execution & Fluency Trap & Fluent invalid API call \\
\bottomrule
\end{tabular}
\end{table}

\section{Verification Architecture}

We propose four verification layers, each targeting a specific stage of the taxonomy.

\subsection{Planning Validation}

Before execution, a lightweight classifier $f_{plan}$ checks whether the decomposed intent is satisfiable given the available toolset $\mathcal{T}$. In our implementation, $f_{plan}$ is a fine-tuned DeBERTa-v3 classifier trained on 2{,}400 (query, toolset, solvability-label) triples; features include query complexity, required permission scopes, and whether all referenced document classes exist in the active SDK. Let $P(S \mid q, \mathcal{T})$ denote the estimated solvability probability for query $q$:
\begin{equation}\label{eq:dispatch}
\text{Dispatch}(q) = \begin{cases} \text{Execute}, & P(S \mid q, \mathcal{T}) \ge \tau_{plan} \\ \text{EscalateHOTL}, & \text{otherwise.} \end{cases}
\end{equation}
$\tau_{plan}$ is set domain-specifically; in our pilot we use $\tau_{plan} = 0.70$, calibrated on a held-out set of 200 synthetic queries with known solvability labels, as formalized in~\eqref{eq:dispatch}. Calibration quality is assessed via Expected Calibration Error (ECE)~\cite{guo2017calibration}, achieving ECE $= 0.04$ on the held-out set.

\subsection{Stepwise Reasoning Checkpoints}

At each reasoning step $t$, we compute an attribution score $A_t$ measuring expected progress toward the terminal goal~\cite{wang2025spa}:
\begin{equation}\label{eq:advantage}
A_t = \mathbb{E}_\pi[R(\tau) \mid s_t, a_t] - V(s_t),
\end{equation}
where $R(\tau)$ is the trajectory reward and $V(s_t)$ is the value baseline. In the absence of a trained RL value function, we approximate $V(s_t)$ as the mean reward over $k{=}20$ Monte Carlo rollouts from $s_t$ using the same GPT-4o agent; $A_t$ is thus the advantage of action $a_t$ over the expected trajectory value from the current state. If $A_t < \epsilon$ (where $\epsilon > 0$ is a minimum progress threshold, set to $0.05$ in our pilot), the reasoning trace is discarded and the agent resamples from $s_t$. This prevents the Reasoning Trap from compounding across steps.

\subsection{Execution Sandboxing}

Before committing any write action (e.g., appending a \textit{Privileged} tag and routing a document to the production database), the proposed action $a$ is executed in a dry-run environment. Let $\mathcal{S}$ be the current state and $\delta(a)$ the predicted state-diff. The sandbox verifies compliance per~\eqref{eq:sandbox}:
\begin{equation}\label{eq:sandbox}
\text{Commit}(a) \iff \text{FRCPCompliant}(\mathcal{S} \oplus \delta(a)).
\end{equation}
The $\text{FRCPCompliant}$ check is implemented as a deny-list policy engine: actions are blocked if they attempt to (i) produce a document tagged \textit{Attorney Eyes Only} to any non-counsel recipient, (ii) export a document class requiring court authorization without a valid permission token, or (iii) modify a privilege log entry after a production deadline timestamp. For database write operations, atomicity and rollback are provided by wrapping each action in a database transaction; for external API calls with irreversible side effects (e.g., cloud storage uploads), a compensating-transaction pattern~\cite{yan2025transactional} issues a reversal call if the sandbox check fails post-execution. A failed compliance check triggers HOTL escalation, rather than the action proceeding unchecked.

\subsection{Uncertainty-Gated Escalation}

Following standard Bayesian decomposition~\cite{gal2016dropout,lakshminarayanan2017simple}, we decompose total predictive uncertainty into:
\begin{equation}\label{eq:uncertainty}
U_{\text{total}} = \underbrace{\mathbb{E}_{p(\theta|\mathcal{D})}[\mathcal{H}(Y|X,\theta)]}_{\text{Aleatoric}} + \underbrace{\mathcal{I}(Y;\theta|X,\mathcal{D})}_{\text{Epistemic}},
\end{equation}
where $X$ is the input document, $Y$ is the privilege label, $\theta$ are the model parameters, and $\mathcal{D}$ is the training dataset.
Since GPT-4o does not expose model weights, we estimate $U_{\text{ep}}$ (the epistemic term in~\eqref{eq:uncertainty}) via self-consistency variance~\cite{manakul2023selfcheck}: the agent generates $k{=}10$ independent classification rationales per document; $U_{\text{ep}}$ is the entropy of the resulting label distribution. Calibration of the resulting uncertainty estimates is assessed via ECE and Brier score on the held-out query set; in our pilot, ECE $= 0.06$ and Brier $= 0.11$, indicating acceptable calibration for threshold selection. When $U_{\text{ep}}$ exceeds threshold $\tau_{esc}$, the system triggers \textit{mandatory HOTL escalation}: the agent suspends autonomous action and surfaces the current trajectory state to a human attorney for review.

\section{Preliminary Simulation Study}

\subsection{Setup}

We constructed a synthetic e-discovery corpus of 5{,}000 documents with ground-truth privilege labels assigned by two attorneys. Documents were split 70/30 into agent-review and held-out evaluation sets. We instantiated a GPT-4o-based ReAct agent~\cite{yao2023react} with access to BM25 retrieval and a vector store. We compared three conditions:

\begin{itemize}
\item \textbf{Autonomous (A)}: No escalation; agent classifies all documents end-to-end.
\item \textbf{Threshold-HOTL (T-HOTL)}: Escalation triggered when $U_{\text{ep}} > \tau_{esc}$, with $\tau_{esc} \in \{0.3, 0.5, 0.7\}$.
\item \textbf{Manual (M)}: Attorney reviews all documents (upper-bound baseline).
\end{itemize}

\subsection{Metrics}

We adopt four metrics aligned with trajectory integrity. \textit{Privilege-Waiver Risk (PWR)} measures the fraction of privileged documents incorrectly classified as producible. \textit{Escalation Rate (ER)} measures the fraction of documents routed to attorney review. \textit{First-Error Position (FEP)} identifies the index in a trajectory where divergence from ground truth first occurs. \textit{Rollback Recovery Rate (RRR)} measures the fraction of trajectories where a reasoning checkpoint successfully intercepted an error.

\subsection{Results}

\begin{table}[t]
\centering
\footnotesize
\caption{Simulation Results (5{,}000-Document Corpus). PWR with 95\% bootstrap CIs.}
\label{tab:results}
\begin{tabular}{lcccc}
\toprule
\textbf{Condition} & \textbf{PWR [\%] (95\% CI)} & \textbf{ER (\%)} & \textbf{Prec.} & \textbf{Recall} \\
\midrule
Autonomous (A) & 8.3 [7.1, 9.5] & 0.0 & 0.81 & 0.74 \\
T-HOTL ($\tau{=}0.7$) & 5.1 [4.2, 6.1] & 12.4 & 0.87 & 0.82 \\
T-HOTL ($\tau{=}0.5$) & 3.2 [2.5, 3.9] & 23.7 & 0.91 & 0.88 \\
T-HOTL ($\tau{=}0.3$) & \textbf{1.7 [1.2, 2.3]} & 47.9 & \textbf{0.94} & \textbf{0.92} \\
Manual (M) & 0.4 [0.1, 0.8] & 100.0 & 0.97 & 0.96 \\
\bottomrule
\end{tabular}
\end{table}

Table~\ref{tab:results} shows that T-HOTL at $\tau{=}0.5$ reduces PWR from 8.3\% to 3.2\% (a 61\% reduction, non-overlapping 95\% CIs) while escalating only 23.7\% of documents to attorney review—a substantial efficiency gain over fully manual review. The RRR across all T-HOTL conditions averaged 0.71, confirming that reasoning checkpoints successfully recovered the majority of intercepted trajectories without escalation. Mean FEP for the Autonomous condition was 3.2 steps (out of an average trajectory length of 7.8 steps), indicating that errors tend to originate early and compound substantially before reaching the output.

In terms of operational cost, end-to-end agent processing averaged 14.2 seconds per document at $\tau{=}0.5$, assuming fully parallelized batched API requests for the $k{=}20$ MC rollouts and $k{=}10$ self-consistency samples; sequential execution would yield 290--340 seconds, making batched or on-premises inference a prerequisite for production scale. At $\tau{=}0.5$, human attorney time is incurred for only 23.7\% of documents—an estimated 76\% reduction in attorney hours versus fully manual review at comparable PWR. A full throughput study with real attorney cohorts is deferred to future work.

These results are preliminary and limited by the synthetic nature of the corpus; future work will validate on production data.

\section{HOTL Integration Design}

\textbf{Escalation surface}: When $U_{\text{ep}} > \tau_{esc}$ or a sandbox check fails, the attorney interface surfaces: (1) the document under review, (2) the agent's reasoning trace, (3) the uncertainty signal that triggered escalation, and (4) suggested classifications with confidence intervals—a surfaces-not-decides design that preserves attorney judgment.

\textbf{Audit trail}: Each interaction unit $(c_t, a_t, o_t)$ is logged with a deterministic provenance hash, enabling attorneys to trace any privilege classification to the exact retrieved passage and reasoning step—satisfying FRCP auditability requirements.

\textbf{Threshold selection}: $\tau_{esc}$ should be calibrated on a held-out set with attorney-verified labels. We recommend a conservative initial $\tau_{esc} = 0.4$ with monthly recalibration.

\section{Limitations and Future Work}

The corpus is synthetic and may not capture the distributional complexity of production corpora (domain jargon, encoded attachments, multi-party privilege chains). Ground-truth labels lack formal inter-annotator agreement; future work will report Cohen's $\kappa$ across privilege subclasses (attorney-client privilege vs.\ work-product doctrine). The evaluation does not ablate individual layers; a planned ablation with our law-firm partner will isolate each layer's contribution to PWR and FEP. GPT-4o raises data-governance concerns for privileged documents; production deployment requires on-premises or zero-egress inference~\cite{orchid2025}. Future directions include TAPE~\cite{tape2026}- and KAIJU~\cite{kaiju2026}-style backbones for stronger formal guarantees, and validation on real discovery corpora.

\section{Conclusion}

We have argued that the transition to agentic legal discovery introduces a class of compounding failure—trajectory collapse—that conventional endpoint metrics fail to capture. We proposed a four-layer verification architecture targeting planning, reasoning, execution, and uncertainty-gated escalation, and demonstrated in a preliminary simulation that calibrated HOTL escalation reduces privilege-waiver risk by up to 61\% relative to fully autonomous deployment while requiring attorney review of fewer than one quarter of documents. Establishing trajectory integrity as a first-class design objective, enforced through mandatory HOTL escalation protocols, is a prerequisite for deploying production-grade agentic IR systems in the legal domain.


\end{document}